\documentclass[preprint,12pt,a4paper]{elsarticle}

\usepackage{amsmath,amsfonts}
\usepackage{amssymb}
\usepackage{hyperref}
\usepackage{listings}
\usepackage{xcolor}
\usepackage{booktabs}
\definecolor{commentcolor}{HTML}{9b9d9b}
\definecolor{codegray}{rgb}{0.5,0.5,0.5}
\definecolor{codepurple}{rgb}{0.58,0,0.82}
\definecolor{codepurple}{rgb}{0.58,0,0.82}
\definecolor{backcolor}{HTML}{ffffff}
\lstdefinestyle{mystyle}{
    backgroundcolor=\color{backcolor},   
    commentstyle=\color{commentcolor},
    keywordstyle=\color{blue},
    numberstyle=\tiny\color{codegray},
    stringstyle=\color{codepurple},
    basicstyle=\linespread{1.0}\ttfamily\scriptsize,
    breakatwhitespace=false,         
    breaklines=true,                 
    captionpos=b,                    
    keepspaces=true,                 
    numbers=left,                    
    numbersep=10pt,                  
    showspaces=false,                
    showstringspaces=false,
    showtabs=false,                  
    tabsize=1,
    frame=lines
}
\usepackage{soul}

\DeclareMathOperator{\sgn}{sgn}
\journal{SoftwareX}

\begin{document}
\renewcommand{\labelenumii}{\arabic{enumi}.\arabic{enumii}}

\begin{frontmatter}
\title{ReModels: Quantile Regression Averaging models}

\author[label1]{Grzegorz Zakrzewski}
\author[label1]{Kacper Skonieczka}
\author[label1]{Mikołaj Małkiński}
\author[label1,label2]{Jacek Mańdziuk}
\address[label1]{Faculty of Mathematics and Information Science, Warsaw University of Technology, Warsaw, Poland}
\address[label2]{Faculty of Computer Science, AGH University of Krakow, Krakow, Poland}

\begin{abstract}
Electricity price forecasts play a crucial role in making key business decisions within the electricity markets. A focal point in this domain are probabilistic predictions, which delineate future price values in a more comprehensive manner than simple point forecasts. The golden standard in probabilistic approaches to predict energy prices is the Quantile Regression Averaging (QRA) method. In this paper, we present a Python package that encompasses the implementation of QRA, along with modifications of this approach that have appeared in the literature over the past few years. The proposed package also facilitates the acquisition and preparation of data related to electricity markets, as well as the evaluation of model predictions. 
\end{abstract}


\end{frontmatter}

\section*{Metadata}
\label{sec:metadata}
A description of code metadata is presented in Table~\ref{tab:metadata}.

\begin{table}[t]
    \begin{tabular}{|l|p{6.5cm}|p{6.5cm}|}
        \hline
        C1 & Current code version & v1.0 \\
        \hline
        C2 & Permanent link to code/repository used for this code version & \url{https://github.com/zakrzewow/remodels} \\
        \hline
        C3  & Permanent link to Reproducible Capsule & \url{/}\\
        \hline
        C4 & Legal Code License & MIT License \\
        \hline
        C5 & Code versioning system used & git \\
        \hline
        C6 & Software code languages, tools, and services used & python \\
        \hline
        C7 & Compilation requirements, operating environments \& dependencies & python 3.8+ \\
        \hline
        C8 & If available Link to developer documentation/manual & \url{https://remodels.readthedocs.io/en/latest/} \\
        \hline
        C9 & Support email for questions & zakrzewski.grzegorz@outlook.com \\
        \hline
    \end{tabular}
    \caption{Code metadata.}
    \label{tab:metadata} 
\end{table}

\section{Motivation and significance}
In recent years, due to growing competition in electricity markets, changes in infrastructure, and the expanding presence of renewable energy sources, predicting electricity prices is an increasingly important issue in planning and operational activities of various entities. Consequently, probabilistic approaches to forecasting energy prices are worth attention, as such predictions offer richer insights into future price dynamics. In certain scenarios the focus lies on forecasting the variability of future price movements rather than mere single-point predictions. Prediction intervals and density forecasts offer supplementary insights into the trajectory of forthcoming prices. In particular, they facilitate a more accurate assessment of future uncertainty and enable the formulation of various strategies that take into account the spectrum of potential outcomes determined by the interval forecast. This enhanced informational depth proves advantageous from multiple perspectives, including business operations, risk management, and stock market decision-making. A more detailed introduction to probabilistic forecasting of energy prices is presented in the review paper~\cite{NOWOTARSKI2018}.

The growing interest in probabilistic energy price forecasting culminated in the international Global Energy Forecasting Competition (GEFCom2014)~\cite{GEFCOM2014}, which yielded significant outcomes. Participants were tasked with developing the most effective model for predicting the 99 percentiles of future prices distribution. Notably, the Quantile Regression Averaging (QRA) method \cite{NOWATORSKI2015} emerged as the preferred approach employed by the two most successful teams. Their spectacular success accelerated the development of the QRA approach in the energy price forecasting field. 

Researchers continue to utilize and refine QRA, what has led to several notable improvements, including QRM~\cite{MARCJASZ2020, SERAFIN2019}, FQRA~\cite{MACIEJOWSKA2016,MACIEJEWSKA2023}, FQRM~\cite{MACIEJOWSKA2016,MACIEJEWSKA2023}, sFQRA~\cite{MACIEJOWSKA2016,MACIEJEWSKA2023}, sFQRM~\cite{MACIEJOWSKA2016,MACIEJEWSKA2023}, LQRA~\cite{UNIEJEWSKI2021}, SQRA~\cite{fernandes2021smoothing,UNIEJEWSKI2023} and SQRM~\cite{fernandes2021smoothing,UNIEJEWSKI2023}. These methods expand the repertoire of techniques available for probabilistic energy price forecasting and are briefly summarized in this section.

To briefly introduce the topic, quantile regression (QR) is employed to estimate the conditional median (or any other quantile) of the response variable. The objective of QR is to express quantiles of the response variable as a linear function of explanatory variables. QR can be considered a special case of linear regression (LR). The primary distinction between QR and LR lies in the cost function being optimized. Quantile regression aims to minimize the following expression:
\begin{equation}\label{eq:qr_final}
     \beta_k = \underset{\beta \in \mathbb{R}^n}{\operatorname{argmin}} \,\mathbb{E} \left[ \rho_k (Y - X \beta) \right]
\end{equation}
where $\beta_k$ are the model parameters for quantile $k \in(0, 1)$, $X$ is the input matrix and $Y$ is the response variable.
As proposed in~\cite{NOWATORSKI2015}, in QRA $X$ comprises predictions of one or many point models and $Y$ corresponds to electricity prices.
The input matrix $X$, which is composed of point predictions of electricity prices, can be obtained using any chosen method. Since electricity prices usually come in hourly frequency, one straightforward method to estimate the future electricity price $P_{d, h}$ for a specific day~$d$ and hour~$h$ is to use known prices $P_{d-1,h}$ from the previous day. A more sophisticated approach would involve training an LR model on historical prices, for example incorporating variables such as prices from previous day $P_{d-1,h}$, two days prior $P_{d-2,h}$, and one week prior $P_{d-7,h}$. Obviously, a target variable in this described LR model is the unknown electricity price $P_{d,h}$. Generated point predictions form a matrix of point predictions. If only one approach was utilized, the resulting point predictions matrix $X$ will have one column. Utilizing various models or training one model with different sets of hyper-parameters will result in a broader matrix, respectively. For each quantile $k$, the parameters of the QR model are estimated separately, utilizing the point predictions matrix. The calculated parameters vector $\beta_k$ is later used to estimate the $k$-th quantile of the future electricity price.

\begin{figure}[t]
    \centering
    \includegraphics[width=\textwidth]{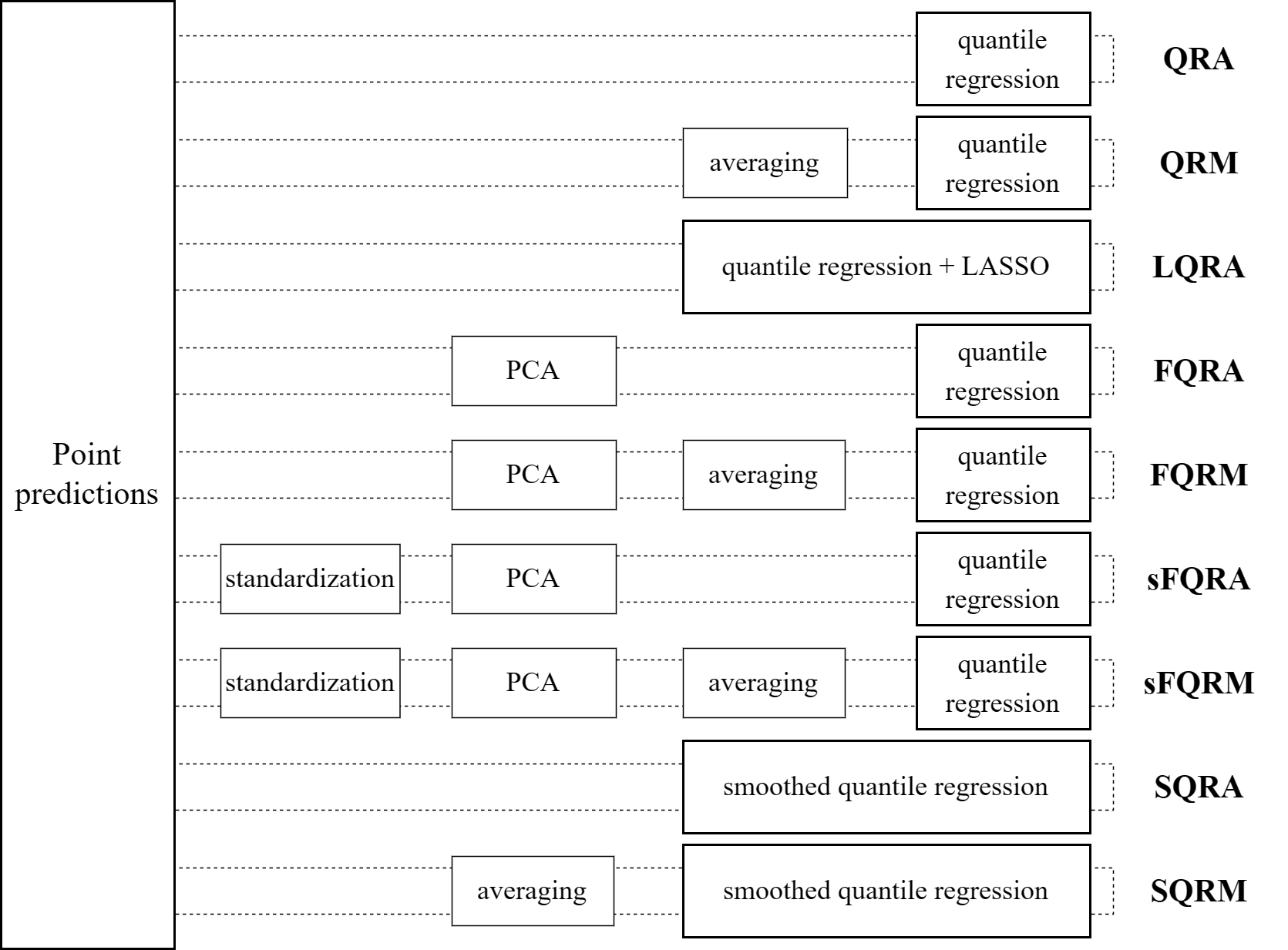}
    \caption{A summary of QRA variants. 
    \textit{Point predictions} block refers to matrix of forecasts of electricity prices obtained in any chosen way, e.g. with autoregressive model trained on historical prices. Multiple series of point predictions can be prepared. \textit{Averaging} involves computing the average of rows in the point predictions matrix. This average serves as the input for the next step. In the \textit{PCA} block, Principal Components of the point predictions matrix are computed, with the number of utilized components being a hyper-parameter. \textit{Standardization} entails scaling the rows in the point predictions matrix w.r.t. their mean and standard deviation.}
    \label{fig:qra_summary}
\end{figure}

As mentioned above, eight main modifications of the base QRA method have 
been proposed. In the first one -- QRM -- point predictions are averaged across rows (the time dimension) before applying QR. LQRA introduces an L1 penalty term to the QR cost function. The primary modification in FQRA and sFQRA involves utilization of Principal Component Analysis (PCA)~\cite{wold1987principal} on the input point predictions matrix, along with optional averaging or standardization. Lastly, the SQRA variant modifies the QR cost function by incorporating kernel density estimation, rendering it convex. A high-level summary of all $9$ QRA variants is provided in Figure \ref{fig:qra_summary}.

Despite achieving state-of-the-art results in energy price forecasting, none of the referenced QRA papers has disclosed their code or the software utilized. Furthermore, no research paper offered detailed implementation instructions. Additionally, while QR is implemented in its most basic form within the \texttt{scikit-learn} and \texttt{statsmodels} Python libraries, the implementation of more advanced QRA variants is missing. In addition, although some public platforms provide data concerning energy prices, the actual datasets used for model training and evaluation aren't shared explicitly, leaving room for inconsistencies in the data acquisition process performed by different research groups. Lastly, the evaluation process of developed methods necessitates the usage of custom metrics that are not available in existing tools, which may lead to discrepancies in the method assessment process. The above reasons motivated us to develop our own package, which encompasses implementations of all $9$ QRA variants and for their assessment on common grounds in terms of both the data and evaluation metrics used.

More specifically, our package, ReModels, comprises several modules designed to enhance energy price forecasting research. Firstly, it enables the users to download public datasets which are commonly used in research papers on the subject, ensuring transparency of the data acquisition process. With the acquired data, users can undertake a series of actions: apply a variance stabilizing transformation (a data preprocessing technique), generate point and probabilistic forecasts using a reference implementation from the above-mentioned set of QRA variants, improving reproducibility of experiments. Finally, the users can evaluate and compare the resulting predictions with dedicated metrics enabling fair and consistent evaluation.

ReModels package offers a comprehensive solution for researchers focused on energy price forecasting using QRA. It enables extensive comparisons between QRA variants, as well as between QRA and other probabilistic energy price forecasting techniques. Additionally, the package serves as a tool in the development of new forecasting methods going beyond the energy price forecasting field.

\section{Software description}

\subsection{Software architecture}
ReModels is a Python package comprising several modules. Its underlying idea is to follow and support a standard schema of the QRA related research:
\begin{enumerate}
    \item dataset acquisition and description;
    \item data preparation - standardization and variance stabilizing transformations;
    \item computation of point forecasts using selected methods;
    \item assessment of the accuracy of point forecasts;
    \item generation of probabilistic forecasts using a chosen QRA variant based on point forecasts from the previous step;
    \item assessment of the accuracy of probabilistic forecasts, often accompanied by comparisons between forecasts of different QRA variants.
\end{enumerate}
ReModels package was designed with these specific steps in mind. Each of them is implemented as a dedicated module. In the following paragraphs, we delve into the properties of these modules.

\subsection{Software functionalities}
\paragraph{Dataset acquisition} The aim of the first module is to grant users access to selected datasets. These datasets typically encompass time series of historical prices in the energy market, along with other market-related variables, such as forecasts of electric energy consumption. With this data, users can construct models for both point and probabilistic predictions. Within this module, a class is implemented to facilitate user access to data from the publicly available ENTSO-E API, presented in the popular form of \texttt{pandas} dataframes. The ENTSO-E Transparency Platform was established to enhance transparency within the European energy market, offering access to current and historical data concerning electricity prices and energy consumption across various countries.

\paragraph{Data preparation} The data preparation module comprises classes that implement various preprocessing techniques, referred to as transformers. Each transformer possesses the capability to fit to the data, execute transformations, and perform inverse transformations. The available transformers allow to scale the data w.r.t. their mean and standard deviation or median and mean absolute deviation, adjust for the daylight saving time, and stabilize the variance of the data. Variance stabilization is a particularly critical step in energy price forecasting, as electricity prices are vulnerable to rapid spikes that produce outliers in the data, which exert a significantly adverse impact on model parameters. Eight different variance stabilizing transformers (VSTs) have been implemented: clipping, clipping with the logarithm function, logistic transform, inverse hyperbolic sine transform, BoxCox transform, polynomial transform, mirror-log transformation, and probability integral transform. VST names and formulas are listed in Table \ref{tab:vst}. All transformations are fully described in \cite{UNIEJEWSKI2017}. The VST methods are applied as follows: point predictions for a specific day and hour ($P_{d,h}$) are scaled, which gives $p_{d,h}$. Then, the selected VST is applied to these scaled predictions ($p_{d,h}$), resulting in $Y_{d,h}$. Subsequently, probabilistic predictions $\hat{Y}_{d,h}$ are calculated based on the transformed values. Finally, an inverse transformation is applied to obtain the final predictions $\hat{P}_{d,h}$, expressed in a real scale and units.
\[
P_{d,h} \xrightarrow{\text{scaling}} p_{d,h} \xrightarrow{\text{VST}} Y_{d,h}  \xrightarrow{\text{prediction}} \hat{Y}_{d,h} \xrightarrow{\text{VST inversion}} \hat{P}_{d,h}
\]
\begin{table}[t]
    \centering
    \small
    \begin{tabular}{l|l}
        \toprule
        Transformation & Formula\\
        \midrule
        $3\sigma$ & $Y_{d,h} = \begin{cases} 3 \cdot \sgn(p_{d,h}) & \text{if } |p_{d,h}| > 3, \\ p_{d,h} & \text{if } |p_{d,h}| \leq 3. \end{cases}$\\
        \midrule
        $3\sigma \log$ & $Y_{d,h} = \begin{cases} \sgn(p_{d,h}) \left( \log(|p_{d,h}| - 2) + 3 \right) & \text{if } |p_{d,h}| > 3, \\ p_{d,h} & \text{if } |p_{d,h}| \leq 3.\end{cases}$\\
        \midrule
        \textit{logistic} & $Y_{d,h} = \left(1 + e^{-p_{d,h}}\right)^{-1}$\\
        \midrule
        \textit{arcsinh} & $Y_{d,h} = \text{arcsinh}(p_{d,h}) \equiv \log\left(p_{d,h} + \sqrt{p_{d,h}^2 + 1}\right)$\\
        \midrule
        \textit{BoxCox} & $Y_{d,h} = \begin{cases} \sgn(p_{d,h})\frac{\left((|p_{d,h}|+1)^{\lambda}-1\right)}{\lambda} & \text{if } \lambda > 0, \\ \log(|p_{d,h}| + 1) & \text{if } \lambda = 0. \end{cases}$\\
        \midrule
        \textit{poly} & $Y_{d,h} = \sgn(p_{d,h})\left[\left(\frac{|p_{d,h}|}{c} + 1\right)^{\lambda} - \left(\frac{1}{c}\right)^{\lambda}\right]^{\frac{1}{\lambda - 1}}$\\
        \midrule
        \textit{mlog} & $Y_{d,h} = \sgn(p_{d,h})\left[\log\left(\frac{|p_{d,h}|}{c} + 1\right) + \log(c)\right]$\\
        \midrule
        \textit{PIT} & $Y_{d,h} = G^{-1}\left(\hat{F}_{P_{d,h}}(P_{d,h})\right)$\\
        \bottomrule
    \end{tabular}
    \caption{Variance Stabilizing Transformations implemented in the ReModels package. \textit{mlog} denotes the mirror logarithm transform, $\lambda$ and $c$ are hyper-parameters specific to \textit{BoxCox}, \textit{poly} and \textit{mlog} methods. Article \cite{UNIEJEWSKI2017} from which these methods were taken does not discuss the relevance of these parameters, but provides their suggested default values. $G^{-1}$ is an inverse of the selected continuous distribution (e.g. normal distribution) and $\hat{F}_{P_{d,h}}$ is an estimate (e.g. empirical cdf) of distribution of electricity prices ${P_{d,h}}$.
    }
    \label{tab:vst}
\end{table}
\paragraph{Point forecasts} The subsequent module focuses on point forecasts. Generally, a point forecast can be obtained using any model or technique, and we aim to avoid imposing restrictions on the package users regarding this matter. The module provides the \texttt{PointModel} class that follows the API of the popular \texttt{Pipeline} class from \texttt{scikit-learn}. The user can instantiate a \texttt{PointModel} from a set of instructions, including scalers, transformers, and, ultimately, the selected model, which must implement \texttt{fit} and \texttt{predict} methods. The prepared pipeline can be executed to calculate point predictions from the provided data. Scalers and transformers preferred by the user are automatically incorporated into the process.

\paragraph{Point forecast evaluation} The fourth module is dedicated to assessing point forecasts. In the field of energy price forecasting, the most commonly utilized measures include Mean Absolute Error (MAE), Root Mean Squared Error (RMSE), and Mean Absolute Percentage Error (MAPE). All the aforementioned metrics are presented in Table \ref{tab:point_metrics} and can be calculated by the point forecast assessment module.
\begin{table}[t]
    \centering
    \small
    \begin{tabular}{l|l}
        \toprule
        Measure & Formula\\
        \midrule
        \textit{MAE} & $\frac{1}{24 N_d} \sum_{d=1}^{N_d} \sum_{h=1}^{24} \left| P_{d,h} - \hat{P}_{d,h} \right|$\\
        \midrule
        \textit{RMSE} & $\sqrt{\frac{1}{24 N_d} \sum_{d=1}^{N_d} \sum_{h=1}^{24} \left( P_{d,h} - \hat{P}_{d,h} \right)^2}$\\
        \midrule
        \textit{MAPE} & $ \frac{1}{24 N_d} \sum_{d=1}^{N_d} \sum_{h=1}^{24} \frac{\left| P_{d,h} - \hat{P}_{d,h} \right|}{\left| P_{d,h} \right|}$\\
        \bottomrule
    \end{tabular}
    \caption{Metrics used to assess point forecasts implemented in the ReModels package. $N_d$ denotes the number of days.}
    \label{tab:point_metrics} 
\end{table}
\paragraph{Probabilistic forecasts} The core of the ReModels package is its probabilistic forecasts generation module. Within this module, all nine QRA variants are implemented as separate classes that inherit from the \texttt{QRA} base class. Each model class includes \texttt{fit} and \texttt{predict} methods, responsible for optimizing model parameters and generating predictions in a manner specific to the respective QRA variant. Some variants, such as FQRA and SQRA, may require additional parameters for model fitting, although reasonable defaults are provided. The organization of models increases cohesion of the probabilistic forecasting module, reduces coupling between the QRA variants implementations, simplifies future maintenance and facilitates package extensibility.

\paragraph{Probabilistic forecast evaluation} The final module allows assessing the accuracy of probabilistic forecasts. It incorporates two metrics: Average Empirical Coverage (AEC), which measures the frequency of calculated prediction intervals containing actual values, and Aggregate Pinball Score (APS), which identifies models with the lowest value of the QR loss function. In addition, the module implements the Kupiec and Christoffersen statistical tests. For convenience, the \texttt{qra.tester} module is provided that unifies common steps of an energy price forecasting pipeline under an intuitive API. These steps include fitting and evaluating the provided QRA model based on a configurable set of calibration and prediction windows.

\section{Illustrative example}
This section offers an illustrative, end-to-end example of how the ReModels package can be used. It involves downloading data related to a selected electricity market, processing the data with various VSTs, obtaining point and probabilistic forecasts, and finally, assessing the accuracy of the calculated forecasts. The full code required to reproduce the example is provided in Listing \ref{lst:example}.

The initial objective, downloading data, can be effortlessly accomplished using the \texttt{remodels.data.EntsoeApi} class. In the example, we utilize a dataset sourced from the German electricity market, containing historical electricity prices and load forecasts from January 5th, 2015 to January 1st, 2017. The user simply needs to provide the start and end dates to the \texttt{get\_day\_ahead\_pricing} and \texttt{get\_forecast\_load} methods of the mentioned class. The header rows of the downloaded data are displayed in Table~\ref{tab:data_example1}.
\begin{table}[t]
    \centering
    \small
    \begin{tabular}{l|r|r}
        % \hline
        \toprule
        datetime & price\_da & quantity\\
        % \hline
        \midrule
        2015-01-04 23:00:00 & 22.34 & 50326.50\\
        2015-01-05 00:00:00 & 17.93 & 48599.50\\
        2015-01-05 01:00:00 & 15.17 & 47364.00\\
        2015-01-05 02:00:00 & 16.38 & 47292.25\\
        2015-01-05 03:00:00 & 17.38 & 48370.25\\
        % \hline
        \bottomrule
    \end{tabular}
    \caption{Five header rows of the downloaded data from the German energy market. \textit{price\_da} is an abbreviation for the day-ahead price [EUR/MWh], which represents the actual price for a given date and time. It is termed ``day-ahead'' because it is typically published the day before its delivery time. \textit{quantity} is a forecast of electricity load [MWh].
    }
    \label{tab:data_example1}
\end{table}

\begin{figure}[t]
    \centering
    \includegraphics[width=\textwidth]{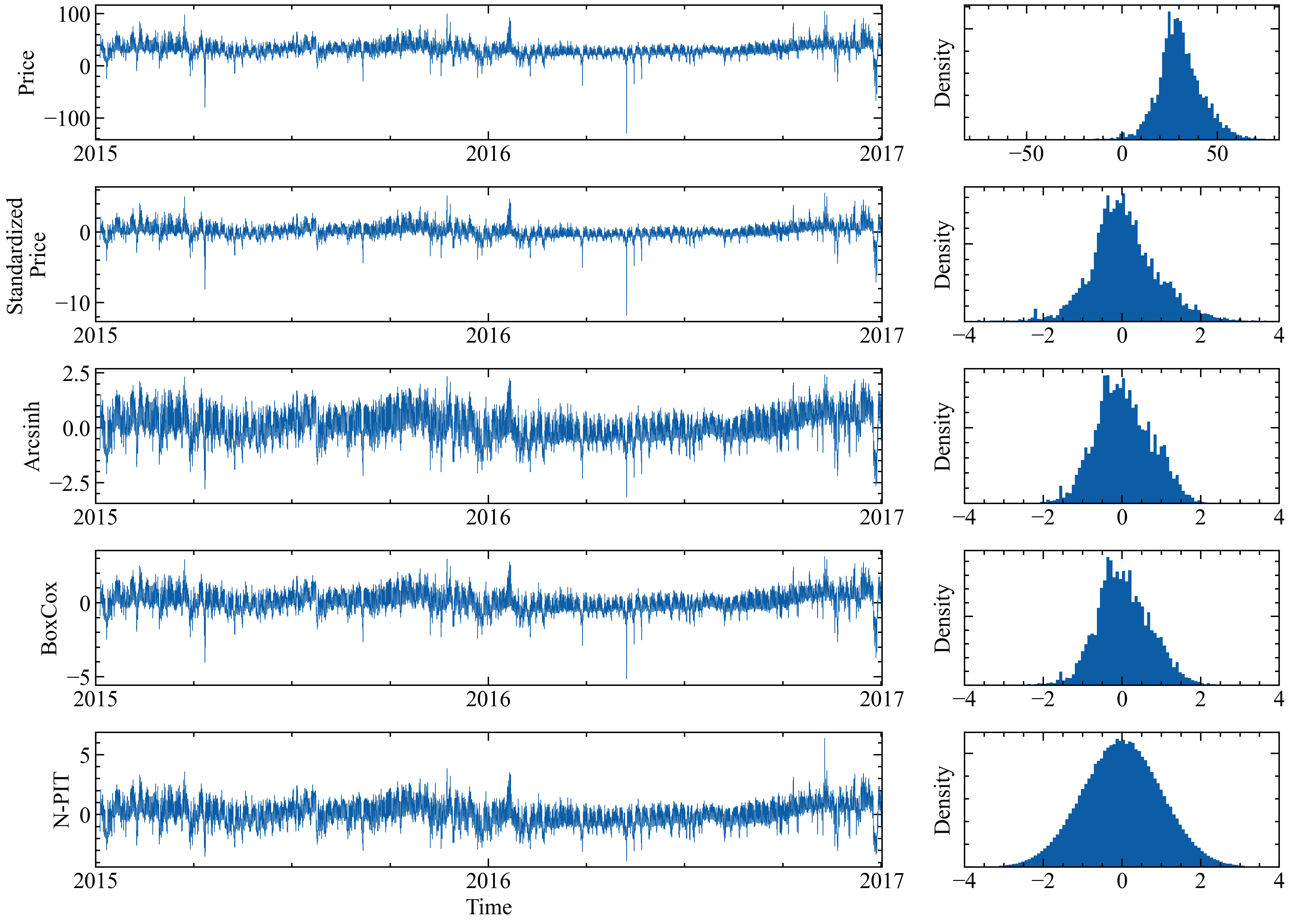}
    \caption{The impact of selected transformations on electricity prices on the German market. The plots present the price over time (left) and the corresponding density (right). 
    } 
    \label{fig:VST_german}
\end{figure}

To process the gathered data, one can utilize a selected transformer from \texttt{remodels.transformers.VSTransformers}. Each transformer class 
implements the \texttt{fit\_transform} method, which handles the necessary processing. Figure~\ref{fig:VST_german} demonstrates the effect of applying selected VSTs on the considered dataset.
The setup of the forecasting model involves defining a sequence of data preprocessing and modeling steps within a pipeline. Steps such as standard scaling, variance stabilization, and the predictive model are encapsulated within \texttt{RePipeline} class. This prepared pipeline is then passed to \texttt{PointModel} class. An essential feature of \texttt{PointModel} is its ability to dynamically retrain on a daily basis during the prediction phase, utilizing the \texttt{predict} method. The model provides the flexibility to generate forecasts over calibration windows of varying lengths. A calibration window represents a subset of data on which the models are retrained. In energy price forecasting, it is a common practice to train the model on a selected range, e.g. 364 days. After model calibration, a prediction is made for the next day. Subsequently, the oldest date in the calibration window is discarded, and a new date is incorporated, effectively shifting the window by one day. Once again, the model is retrained, and prediction is made for the subsequent date. \texttt{PointModel} class moves the calibration window of the selected length across the data.
The \texttt{pointModel.summary()} method facilitates an in-depth examination of performance metrics across various calibration windows, akin to the example shown in Table~\ref{tab:model_performance}.
\begin{table}[t]
    \centering
    \small
    \begin{tabular}{l|ccccc}
        % \hline
        \toprule
        Calibration Window & MAE & MSE & RMSE & MAPE & $R^2$ \\
        % \hline
        \midrule
        182 days & 10.60 & 255.07 & 15.97 & 38.31 & 0.616 \\
        364 days & 10.57 & 242.09 & 15.55 & 56.93 & 0.636 \\
        728 days & 11.03 & 263.64 & 16.23 & 68.70 & 0.603 \\
        % \hline
        \bottomrule
    \end{tabular}
    \caption{Summary of performance metrics obtained from different calibration windows using the \texttt{pointModel.summary()} method for prices from the German market.}
    \label{tab:model_performance}
\end{table}
The \texttt{remodels.qra} module provides implementation of a wide set of QRA variants. Their application to the selected dataset involves using the \texttt{fit} method to train the model on the data and the \texttt{predict} method to forecast the specified quantiles. This two-step process - fitting and predicting - enables detailed probabilistic analysis across a range of quantiles, enhancing forecasting precision and reliability.

The \texttt{remodels.qra.tester.QR\_Tester} class enables comprehensive testing across designated calibration and prediction windows. In particular, \texttt{QR\_Tester} offers computing key evaluation metrics, such as the AEC or APS. For illustrative purposes, Table~\ref{tab:qra_performance} presents AEC for several QRA variants and expected coverage thresholds, providing a snapshot of the models' performance.
\begin{table}[ht]
    \centering
    \small
    \begin{tabular}{l|c|c|c}
        % \hline
        \toprule
        Expected Coverage& 50\% & 70\% & 90\% \\
        % \hline
        \midrule
        QRA  & 45.17 & 63.51 & 84.86 \\
        QRM  & 45.87 & 65.05 & 85.77 \\
        SQRA & 50.96 & 69.81 & 88.36 \\
        SQRM & 52.30 & 70.71 & 88.87 \\
        % \hline
        \bottomrule
    \end{tabular}
    \caption{Average Empirical Coverage (AEC) is calculated using probabilistic predictions from selected four QRA variants across expected coverage levels. AEC is a measure of coverage of prediction intervals obtained with quantile regression. To obtain a prediction interval, two quantiles must be selected. For instance, the 25th and 75th quantiles create a 50\% prediction interval. By considering all prediction intervals of this length, we anticipate that 50\% of actual price values will fall within these intervals - this is our expected coverage. The frequency of calculated prediction intervals containing actual values is typically either lower or higher, but achieving closer proximity to the expected coverage indicates better performance. In the literature on QRA, prediction intervals with expected coverage of 50\%, 70\%, and 90\% are commonly considered.}
    \label{tab:qra_performance}
\end{table}
\begin{lstlisting}[caption={The code required to reproduce the illustrative example.}, label={lst:example}, language=Python]
# data downloading
import datetime as dt
from remodels.data.EntsoeApi import EntsoeApi

start_date = dt.date(2015, 1, 1)
end_date = dt.date(2017, 1, 1)
security_token = "..."  # free token from https://transparency.entsoe.eu/
entsoe_api = EntsoeApi(security_token)
prices = entsoe_api.get_day_ahead_pricing(
    start_date,
    end_date,
    "10Y1001A1001A63L",  # Germany domain code in ENTSO-E
    resolution_preference=60,  # resolution in minutes
)
forecast_load = entsoe_api.get_forecast_load(
    start_date, 
    end_date, 
    "10Y1001A1001A63L"
)
germany_data = prices.join(forecast_load)

# VST & point predictions
from remodels.pipelines.RePipeline import RePipeline
from remodels.pointsModels import PointModel
from remodels.transformers import StandardizingScaler
from remodels.transformers.VSTransformers import ArcsinhScaler
from sklearn.linear_model import LinearRegression

pipeline = RePipeline(
    [
        ("standardScaler", StandardizingScaler()),
        ("vstScaler", ArcsinhScaler()),
        ("linearRegression", LinearRegression()),
    ]
)
pointModel = PointModel(pipeline=pipeline)
pointModel.fit(germany_data, dt.date(2016, 12, 1), dt.date(2016, 12, 31))
point_predictions = pointModel.predict(calibration_window=182)

# point predictions metrics
pointModel.summary()

# probabilistic predictions
from remodels.qra import QRA
from remodels.qra.tester import QR_Tester

# actual prices - target variable in QRA model
price_da = point_predictions.join(germany_data)["price_da"]
# selected QRA variant
qra_model = QRA(fit_intercept=True)

results = QR_Tester(
    calibration_window=72,
    qr_model=qra_model
).fit_predict(point_predictions, price_da)

# probabilistic predictions metrics
# alpha=50 is excpected coverage of prediction intervals
results.aec(alpha=50)  # Average Empirical Coverage
results.kupiec_test(alpha=50, significance_level=0.05) 
results.christoffersen_test(alpha=50, significance_level=0.05) 
results.aps()  # Aggregate Pinball Score
\end{lstlisting}

\section{Impact}
ReModels addresses the reproducibility challenge in energy price forecasting emphasized by Lago et al.~\cite{LAGO}. By providing a transparent approach to data acquisition, method evaluation and result analysis, the package enhances fairness and consistency of conducted experiments. We expect the package will lead to an accelerated development in energy price forecasting research by simplifying all critical steps of the modelling pipeline.

The package has been applied internally in our research group to perform a fundamental comparison of implemented QRA methods across several electricity markets and date ranges. The choice of markets and date ranges used for method evaluation differs from paper to paper, which makes comparison of new results with prior work difficult. In contrast, the open-source nature of ReModels facilitates gradual extension of published results to new evaluation settings (e.g. new market or date range).

We’ve utilised the package to study the impact of incorporating features concerning renewable energy sources for probabilistic price forecasting. To this end, we injected new input features into the point prediction model, while the remaining aspects of the modelling pipeline remained unchanged, thus significantly reducing the time needed to validate formulated hypotheses. We expect that similar benefits will be experienced by the users of the package, including researchers and engineers that aim to explore and assess new modelling techniques, as well as analysts and decision-makers interested in applying an existing toolkit to new data.

Although we focus on the energy price forecasting problem, the applicability of the proposed package goes beyond this domain, as QRA methods are widely adopted also in other fields, e.g. load \cite{liu2015probabilistic,zhang2018parallel}, wind power \cite{zhang2016deterministic}, or nodal voltage forecasting \cite{wang2021short}. Moreover, QRA serves as a benchmark for other probabilistic energy price forecasting methods, such as distributional neural networks \cite{MARCJASZ2023106843,barunik2023learning} or conformal prediction intervals \cite{KATH2021777}. QRA is also compared to its more distant cousins, such as Quantile Regression Random Forest \cite{cornell2024probabilistic,8356127},  Quantile Regression Gradient Boosting \cite{8356127,7423794} or Quantile Regression Neural Network \cite{gan2018embedding,8419220}, which are also utilized in fields like price and electricity consumption forecasting.

Due to its modular construction, ReModels provides a set of reusable components for tasks requiring probabilistic predictions. Coupled with the compatibility with the scikit-learn package, ReModels provides a general prediction toolkit applicable across various fields. 
In future work, we plan to enhance ReModels capabilities by implementing additional models. These models include Quantile Regression Forest \cite{meinshausen2006quantile} and Quantile Regression Neural Network \cite{taylor2000quantile}. Though these models are not linear, they are related to QRA and can be utilized for probabilistic energy price forecasting.

\section{Conclusions}
ReModels introduces a significant leap in electricity price forecasting with its comprehensive implementation of QRA and its recent enhancements. This Python package provides a streamlined approach to probabilistic forecasting, facilitating the acquisition and preprocessing of data, as well as the evaluation of forecasts. By incorporating various QRA variants, ReModels enables users to explore and compare different forecasting methodologies, enriching the field of energy market analysis. The package's design and functionality are geared towards improving forecast accuracy and aiding decision-making processes in the energy sector.
\bibliographystyle{elsarticle-num} 
\bibliography{softwarex-osp}

\begin{thebibliography}{10}
\expandafter\ifx\csname url\endcsname\relax
  \def\url#1{\texttt{#1}}\fi
\expandafter\ifx\csname urlprefix\endcsname\relax\def\urlprefix{URL }\fi
\expandafter\ifx\csname href\endcsname\relax
  \def\href#1#2{#2} \def\path#1{#1}\fi

\bibitem{NOWOTARSKI2018}
J.~Nowotarski, R.~Weron, Recent advances in electricity price forecasting: A
  review of probabilistic forecasting, Renewable and Sustainable Energy Reviews
  81 (2018) 1548--1568.
\newblock \href {https://doi.org/https://doi.org/10.1016/j.rser.2017.05.234}
  {\path{doi:https://doi.org/10.1016/j.rser.2017.05.234}}.

\bibitem{GEFCOM2014}
T.~Hong, P.~Pinson, S.~Fan, H.~Zareipour, A.~Troccoli, R.~J. Hyndman,
  Probabilistic energy forecasting: Global energy forecasting competition 2014
  and beyond, International Journal of Forecasting 32~(3) (2016) 896--913.
\newblock \href
  {https://doi.org/https://doi.org/10.1016/j.ijforecast.2016.02.001}
  {\path{doi:https://doi.org/10.1016/j.ijforecast.2016.02.001}}.

\bibitem{NOWATORSKI2015}
J.~Nowotarski, R.~Weron, Computing electricity spot price prediction intervals
  using quantile regression and forecast averaging, Computational Statistics 30
  (2015) 791--803.
\newblock \href {https://doi.org/https://doi.org/10.1007/s00180-014-0523-0}
  {\path{doi:https://doi.org/10.1007/s00180-014-0523-0}}.

\bibitem{MARCJASZ2020}
G.~Marcjasz, B.~Uniejewski, R.~Weron, Probabilistic electricity price
  forecasting with narx networks: Combine point or probabilistic forecasts?,
  International Journal of Forecasting 36~(2) (2020) 466--479.
\newblock \href
  {https://doi.org/https://doi.org/10.1016/j.ijforecast.2019.07.002}
  {\path{doi:https://doi.org/10.1016/j.ijforecast.2019.07.002}}.

\bibitem{SERAFIN2019}
T.~Serafin, B.~Uniejewski, R.~Weron, Averaging predictive distributions across
  calibration windows for day-ahead electricity price forecasting, Energies
  12~(13) (2019) 2561.
\newblock \href {https://doi.org/https://doi.org/10.3390/en12132561}
  {\path{doi:https://doi.org/10.3390/en12132561}}.

\bibitem{MACIEJOWSKA2016}
K.~Maciejowska, J.~Nowotarski, R.~Weron, Probabilistic forecasting of
  electricity spot prices using factor quantile regression averaging,
  International Journal of Forecasting 32~(3) (2016) 957--965.
\newblock \href
  {https://doi.org/https://doi.org/10.1016/j.ijforecast.2014.12.004}
  {\path{doi:https://doi.org/10.1016/j.ijforecast.2014.12.004}}.

\bibitem{MACIEJEWSKA2023}
K.~Maciejowska, T.~Serafin, B.~Uniejewski, Probabilistic forecasting with
  factor quantile regression: Application to electricity trading, arXiv
  preprint arXiv:2303.08565 (2023).
\newblock \href {https://doi.org/https://doi.org/10.48550/arXiv.2303.08565}
  {\path{doi:https://doi.org/10.48550/arXiv.2303.08565}}.

\bibitem{UNIEJEWSKI2021}
B.~Uniejewski, R.~Weron, Regularized quantile regression averaging for
  probabilistic electricity price forecasting, Energy Economics 95 (2021)
  105121.
\newblock \href {https://doi.org/https://doi.org/10.1016/j.eneco.2021.105121}
  {\path{doi:https://doi.org/10.1016/j.eneco.2021.105121}}.

\bibitem{fernandes2021smoothing}
M.~Fernandes, E.~Guerre, E.~Horta, Smoothing quantile regressions, Journal of
  Business \& Economic Statistics 39~(1) (2021) 338--357.
\newblock \href {https://doi.org/https://doi.org/10.1080/07350015.2019.1660177}
  {\path{doi:https://doi.org/10.1080/07350015.2019.1660177}}.

\bibitem{UNIEJEWSKI2023}
B.~Uniejewski, Smoothing quantile regression averaging: A new approach to
  probabilistic forecasting of electricity prices, arXiv preprint
  arXiv:2302.00411 (2023).
\newblock \href {https://doi.org/https://doi.org/10.48550/arXiv.2302.00411}
  {\path{doi:https://doi.org/10.48550/arXiv.2302.00411}}.

\bibitem{wold1987principal}
S.~Wold, K.~Esbensen, P.~Geladi, Principal component analysis, Chemometrics and
  intelligent laboratory systems 2~(1-3) (1987) 37--52.
\newblock \href {https://doi.org/https://doi.org/10.1016/0169-7439(87)80084-9}
  {\path{doi:https://doi.org/10.1016/0169-7439(87)80084-9}}.

\bibitem{UNIEJEWSKI2017}
B.~Uniejewski, R.~Weron, F.~Ziel, Variance stabilizing transformations for
  electricity spot price forecasting, IEEE Transactions on Power Systems 33~(2)
  (2017) 2219--2229.
\newblock \href {https://doi.org/https://doi.org/10.1109/TPWRS.2017.2734563}
  {\path{doi:https://doi.org/10.1109/TPWRS.2017.2734563}}.

\bibitem{LAGO}
J.~Lago, G.~Marcjasz, B.~{De Schutter}, R.~Weron, Forecasting day-ahead
  electricity prices: A review of state-of-the-art algorithms, best practices
  and an open-access benchmark, Applied Energy 293 (2021) 116983.
\newblock \href
  {https://doi.org/https://doi.org/10.1016/j.apenergy.2021.116983}
  {\path{doi:https://doi.org/10.1016/j.apenergy.2021.116983}}.

\bibitem{liu2015probabilistic}
B.~Liu, J.~Nowotarski, T.~Hong, R.~Weron, Probabilistic load forecasting via
  quantile regression averaging on sister forecasts, IEEE Transactions on Smart
  Grid 8~(2) (2015) 730--737.
\newblock \href {https://doi.org/https://doi.org/10.1109/TSG.2015.2437877}
  {\path{doi:https://doi.org/10.1109/TSG.2015.2437877}}.

\bibitem{zhang2018parallel}
W.~Zhang, H.~Quan, D.~Srinivasan, Parallel and reliable probabilistic load
  forecasting via quantile regression forest and quantile determination, Energy
  160 (2018) 810--819.
\newblock \href {https://doi.org/https://doi.org/10.1016/j.energy.2018.07.019}
  {\path{doi:https://doi.org/10.1016/j.energy.2018.07.019}}.

\bibitem{zhang2016deterministic}
Y.~Zhang, K.~Liu, L.~Qin, X.~An, Deterministic and probabilistic interval
  prediction for short-term wind power generation based on variational mode
  decomposition and machine learning methods, Energy Conversion and Management
  112 (2016) 208--219.
\newblock \href
  {https://doi.org/https://doi.org/10.1016/j.enconman.2016.01.023}
  {\path{doi:https://doi.org/10.1016/j.enconman.2016.01.023}}.

\bibitem{wang2021short}
Y.~Wang, L.~Von~Krannichfeldt, T.~Zufferey, J.-F. Toubeau, Short-term nodal
  voltage forecasting for power distribution grids: An ensemble learning
  approach, Applied Energy 304 (2021) 117880.
\newblock \href
  {https://doi.org/https://doi.org/10.1016/j.apenergy.2021.117880}
  {\path{doi:https://doi.org/10.1016/j.apenergy.2021.117880}}.

\bibitem{MARCJASZ2023106843}
G.~Marcjasz, M.~Narajewski, R.~Weron, F.~Ziel, Distributional neural networks
  for electricity price forecasting, Energy Economics 125 (2023) 106843.
\newblock \href {https://doi.org/https://doi.org/10.1016/j.eneco.2023.106843}
  {\path{doi:https://doi.org/10.1016/j.eneco.2023.106843}}.

\bibitem{barunik2023learning}
J.~Barun{\'\i}k, L.~Hanus, Learning probability distributions of day-ahead
  electricity prices, Available at SSRN (2023).
\newblock \href {https://doi.org/https://dx.doi.org/10.2139/ssrn.4592411}
  {\path{doi:https://dx.doi.org/10.2139/ssrn.4592411}}.

\bibitem{KATH2021777}
C.~Kath, F.~Ziel, Conformal prediction interval estimation and applications to
  day-ahead and intraday power markets, International Journal of Forecasting
  37~(2) (2021) 777--799.
\newblock \href
  {https://doi.org/https://doi.org/10.1016/j.ijforecast.2020.09.006}
  {\path{doi:https://doi.org/10.1016/j.ijforecast.2020.09.006}}.

\bibitem{cornell2024probabilistic}
C.~Cornell, N.~T. Dinh, S.~A. Pourmousavi, A probabilistic forecast methodology
  for volatile electricity prices in the australian national electricity
  market, International Journal of Forecasting (2024).
\newblock \href {https://doi.org/https://doi.org/10.48550/arXiv.2311.07289}
  {\path{doi:https://doi.org/10.48550/arXiv.2311.07289}}.

\bibitem{8356127}
Y.~Wang, N.~Zhang, Y.~Tan, T.~Hong, D.~S. Kirschen, C.~Kang, Combining
  probabilistic load forecasts, IEEE Transactions on Smart Grid 10~(4) (2019)
  3664--3674.
\newblock \href {https://doi.org/10.1109/TSG.2018.2833869}
  {\path{doi:10.1109/TSG.2018.2833869}}.

\bibitem{7423794}
S.~Ben~Taieb, R.~Huser, R.~J. Hyndman, M.~G. Genton, Forecasting uncertainty in
  electricity smart meter data by boosting additive quantile regression, IEEE
  Transactions on Smart Grid 7~(5) (2016) 2448--2455.
\newblock \href {https://doi.org/10.1109/TSG.2016.2527820}
  {\path{doi:10.1109/TSG.2016.2527820}}.

\bibitem{gan2018embedding}
D.~Gan, Y.~Wang, S.~Yang, C.~Kang, Embedding based quantile regression neural
  network for probabilistic load forecasting, Journal of Modern Power Systems
  and Clean Energy 6~(2) (2018) 244--254.
\newblock \href {https://doi.org/https://doi.org/10.1007/s40565-018-0380-x}
  {\path{doi:https://doi.org/10.1007/s40565-018-0380-x}}.

\bibitem{8419220}
W.~Zhang, H.~Quan, D.~Srinivasan, An improved quantile regression neural
  network for probabilistic load forecasting, IEEE Transactions on Smart Grid
  10~(4) (2019) 4425--4434.
\newblock \href {https://doi.org/10.1109/TSG.2018.2859749}
  {\path{doi:10.1109/TSG.2018.2859749}}.

\bibitem{meinshausen2006quantile}
N.~Meinshausen, G.~Ridgeway, Quantile regression forests., Journal of machine
  learning research 7~(6) (2006).

\bibitem{taylor2000quantile}
J.~W. Taylor, A quantile regression neural network approach to estimating the
  conditional density of multiperiod returns, Journal of forecasting 19~(4)
  (2000) 299--311.

\end{thebibliography}

\end{document}